\documentclass[journal]{IEEEtran}
\usepackage{graphicx}
\usepackage{hyperref}
\usepackage{url}
\usepackage{textcomp} 
\usepackage{xcolor}
\usepackage{ulem} 
\usepackage{tabularx}
\usepackage{caption}
\usepackage{subcaption}
\usepackage{lscape}
\usepackage{float}
\usepackage{tikz}
\usepackage{dblfloatfix} 
\usepackage{grffile} 
\usepackage{float} 
\usepackage{xspace} 
\hypersetup{
    colorlinks = true,
    linkcolor=red,
    citecolor=red,
    urlcolor=blue,
    linktocpage 
}

\newcommand{\banosdataset}[0]{Banos \textit{et al}\xspace}
\newcommand{\recofitdataset}[0]{Recofit\xspace}
\newcommand{\hoeffdingtree}[0]{HT\xspace}
\newcommand{\mondrianforest}[0]{MF\xspace}

\newcommand{\mondriantrees}[0]{Mondrian trees\xspace}
\newcommand{\naivebayes}[0]{NB\xspace}
\newcommand{\FNN}[0]{FNN\xspace}
\newcommand{\mcnn}[0]{MCNN\xspace}
\newcommand{\mcnns}[0]{MCNNs\xspace}
\newcommand{\har}[0]{HAR\xspace}
\newcommand{\knn}[0]{\textit{k}-NN\xspace}
\newcommand{\streamdmcpp}[0]{StreamDM-C++\xspace}

\title{A benchmark of data stream classification
for human activity recognition
on connected objects}
\author{Martin Khannouz and Tristan Glatard\\Department of Computer Science and Software Engineering\\
Concordia University, Montreal, Canada}

\begin{document}
\maketitle

\begin{abstract}
This paper evaluates data stream classifiers from the
perspective of connected devices, focusing on the use case of
\har. We measure both classification performance and
resource consumption (runtime, memory, and power) of five usual stream
classification algorithms, implemented in a consistent library, and applied
to two real human activity datasets and to three synthetic datasets.
Regarding classification performance, results show an overall superiority
of the \hoeffdingtree, the \mondrianforest, and the \naivebayes classifiers
over the \FNN and the Micro Cluster Nearest Neighbor (\mcnn) classifiers on
4 datasets out of 6, including the real ones. In addition, the
\hoeffdingtree, and to some extent \mcnn, are the only classifiers that can
recover from a concept drift. Overall, the three leading classifiers still
perform substantially lower than an offline classifier on the real
datasets. Regarding resource consumption, the \hoeffdingtree and the
\mondrianforest are the most memory intensive and have the longest runtime, however,
no difference in power consumption is found between classifiers. We
conclude that stream learning for \har on connected objects is challenged
by two factors which could lead to interesting future work: a high
memory consumption and low F1 scores overall.
\end{abstract}



\section{Introduction}
\label{sec:introduction}

Internet of Things applications may adopt a centralized model, where connected
objects transfer data to servers with adequate computing capabilities, or a
decentralized model, where data is analyzed directly on the connected objects or
on nearby devices. While the decentralized model limits network transmission,
increases battery life~\cite{sensor-network-survey, sensor-energy-model}, and
reduces data privacy risks, it also raises important processing challenges due
to the modest computing capacity of connected objects. Indeed, it is not
uncommon for wearable devices and other smart objects to include a processing
memory of less than 100~KB, little to no storage memory, a slow CPU, and no
operating system. With multiple sensors producing data at a high frequency,
typically 50~Hz to 800~Hz, processing speed and memory consumption become
critical properties of data analyses. 

Data stream processing algorithms are precisely designed to analyze virtually
infinite sequences of data elements with reduced amounts of working memory.
Several classes of stream processing algorithms were developed in the past
decades, such as filtering, counting, or sampling
algorithms~\cite{kejariwal2015}.  These algorithms must follow multiple
constraints such as a constant processing time per data element, or a constant space
complexity~\cite{issues_learning_from_stream}.  Our study focuses on supervised
classification, a key component of contemporary data models.

We evaluate supervised data stream classifiers from the point of view of
connected objects, with a particular focus on Human Activity Recognition~(\har). The main motivating use case
is that of wearable sensors measuring 3D acceleration and orientation at
different locations on the human body, from which activities such as gym
exercises have to be predicted. A previously untrained supervised classifier is
deployed directly on the wearables or on a nearby object, perhaps a watch, and
aggregates the data, learns a data model, predicts the current activity, and
episodically receives true labels from the human subject. Our main question is
to determine whether on-chip classification is feasible in this context. 

We evaluate existing classifiers from the complementary angles of (1)
classification performance, including in the presence of concept drift, and
(2) resource consumption, including memory usage and classification time
per element (latency). We consider six datasets in our benchmark, including
three that are derived from the two most popular open datasets used for \har, and
three simulated datasets.

Compared to the previous works reviewed in Section~\ref{sec:related-work}, the contributions of our paper are the following:
\begin{itemize}
    \item We compare the most popular data stream classifiers on the specific case of \har;
    \item We provide quantitative measurements of memory and power consumption, as well as runtime;
    \item We implement data stream classifiers in a consistent software library meant for deployment on embedded systems.
\end{itemize} 
The subsequent sections present the materials, methods, and results of our benchmark.


\section{Related Work}

\label{sec:related-work}

To the best of our knowledge, no previous study focused on the comparison
of data stream classifiers for \har in the context of limited memory and
available runtime that characterizes connected objects.

\subsection{Comparisons of data stream classifiers}

Data stream classifiers were compared mostly using synthetic datasets
or real but general-purpose ones (Electrical, CoverType, Poker), which is 
not representative of our use case. In addition, memory and runtime usage 
are rarely reported, with the notable exception of~\cite{StreamDM-CPP}.

The work
in~\cite{prasad2016stream} reviews an extensive list of classifiers for
data streams, comparing the Hoeffding Tree~(\hoeffdingtree), the Naïve Bayes~(\naivebayes), and the k-nearest neighbor~(\knn)
online classifiers. The paper reports an accuracy of 92 for online \knn, 80
for the \hoeffdingtree, and 60 for \naivebayes. The study is limited to a
single dataset (CoverType). 

The work in ~\cite{kaur2020} compares four classifiers (Bayesnet,
\hoeffdingtree, \naivebayes, and Decision Stump) using synthetic datasets.
It reports a similar accuracy of 90 for the Bayesnet, the
\hoeffdingtree, and \naivebayes classifiers, while the Decision Stump one
only reaches 65. Regarding runtimes, Bayesnet is found to be four times
slower than the \hoeffdingtree which is itself three times slower than
\naivebayes and Decision Stump.

The work in~\cite{priya2020comprehensive} compares ensemble classifiers on
imbalanced data streams with concept drifts, using two real datasets
(Electrical, Intrusion), synthetic datasets, and six classifiers, including the
\naivebayes and the \hoeffdingtree ones. The \hoeffdingtree is found to be
the second most accurate classifier after the Accuracy Updated Ensemble.

The authors in~\cite{lopes2020evaluating} have analyzed the resource trade-offs
of six online decision trees applied to edge computing. Their results showed that the
Very Fast Decision Tree and the Strict Very Fast Decision Tree were the most
energy-friendly, the latter having the smallest memory footprint. On the
other hand, the best predictive performances were obtained in combination with
OLBoost. In particular, the paper reports an accuracy of 89.6\% on the Electrical
dataset, and 83.2\% on an Hyperplane dataset.

Finally, the work in~\cite{StreamDM-CPP} describes the architecture of
\streamdmcpp and presents an extensive benchmark of tree-based classifiers,
covering runtime, memory, and accuracy. Compared to other tree-based
classifiers, the \hoeffdingtree classifier is found to have the smallest
memory footprint while the Hoeffding Adaptive Tree classifier is found to be
the most accurate on most of the datasets. 

\subsection{Offline and data stream classifiers for \har}

Several studies evaluated classifiers for \har in an offline (non data stream)
setting. In particular, the work in~\cite{Janidarmian_2017} compared 293
classifiers using various sensor placements and window sizes, concluding on the
superiority of k nearest neighbors (\knn) and pointing out a trade-off between
runtime and classification performance. Resource consumption, including memory
and runtime, was also studied for offline classifiers, such as
in~\cite{memory_consumption_machine_learning} for the particular case of the R
programming language.

In addition, the work in \cite{ugulino2012} achieved an offline accuracy of
99.4\% on a five-class dataset of \har. The authors used AdaBoost, an
ensemble method, with ten offline decision trees. The work in
\cite{ahmed2019smartphone} proposes a Support Vector Machine enhanced with
feature selection. Using smartphone data, the model showed above 90\%
accuracy on day-to-day human activities. Finally, the work in
\cite{san2018robust} applies three offline classifiers to smartphone and
smartwatch human activity data. Results show that Convolutional Neural Network and Random
Forest achieve F1 score of 0.98 with smartwatches and 0.99 with
smartphones.

In a data stream (online) setting, the work in~\cite{omid_2019} presents a wearable
system capable of running pre-trained classifiers on the chip with high classification
accuracy. It shows the superiority of the proposed Feedforward Neural
Network~(\FNN) over \knn.




\section{Materials and Methods}
We evaluate 5 classifiers implemented in either \streamdmcpp~\cite{StreamDM-CPP}
or OrpailleCC~\cite{OrpailleCC}.  \streamdmcpp is a C++ implementation of
StreamDM~\cite{StreamDM}, a software to mine big data streams using
\href{https://spark.apache.org/streaming/}{Apache Spark Streaming}. \streamdmcpp
is usually faster than StreamDM in single-core environments, due to the overhead
induced by Spark.

OrpailleCC is a collection of data stream algorithms developed for embedded
devices. The key functions, such as random number generation or memory
allocation, are parametrizable through class templates and can thus be
customized on a given execution platform.  OrpailleCC is not limited to
classification algorithms, it implements other data stream algorithms such as
the Cuckoo filter~\cite{cuckoo} or a multi-dimensional extension of the
Lightweight Temporal Compression~\cite{multi-ltc}. We extended it with a few
classifiers for the purpose of this benchmark.

This benchmark includes five popular classification algorithms.  The
Mondrian forest~(\mondrianforest)~\cite{mondrian2014} builds decision trees without immediate need
for labels, which is useful in situations where labels are
delayed~\cite{stream_learning_review}.  The Micro-Cluster Nearest
Neighbors~\cite{mc-nn} is a compressed version of the k-nearest neighbor~(\knn)
that was shown to be among the most accurate classifiers for \har from wearable
sensors~\cite{Janidarmian_2017}. The \naivebayes~\cite{naive_bayes} classifier
builds a table of attribute occurrence to estimate class likelihoods.  The
\hoeffdingtree~\cite{VFDT} builds a decision tree using the Hoeffding Bound to
estimate when the best split is found.  Finally, Neural Network classifiers have
become popular by reaching or even exceeding human performance in many fields
such as image recognition or game playing. We use a Feedforward Neural
Network~(\FNN) with one hidden layer,
as described in~\cite{omid_2019} for the recognition of fitness activities on a
low-power platform.

The remainder of this section details the datasets, classifiers, evaluation
metrics and parameters used in our benchmark.

\subsection{Datasets}
\label{sec:method-dataset}
\subsubsection{\banosdataset}
The \banosdataset dataset~\cite{Banos_2014} is a
human activity dataset with 17 participants
and 9 sensors per
participant\footnote{\banosdataset dataset
available
\href{https://archive.ics.uci.edu/ml/datasets/REALDISP+Activity+Recognition+Dataset\#:\~:text=The\%20REALDISP\%20(REAListic\%20sensor\%20DISPlacement,\%2Dplacement\%20and\%20induced\%2Ddisplacement.}{here}.}. Each sensor samples a 3D
acceleration, gyroscope, and magnetic field, as
well as the orientation in a quaternion format,
producing a total of 13 values.  Sensors are
sampled at 50~Hz, and each sample is associated
with one of 33 activities. In addition to the 33
activities, an extra activity labeled 0 indicates
no specific activity.

We pre-process the \banosdataset dataset using
non-overlapping windows of one second (50
samples), and using only the 6 axes (acceleration
and gyroscope)
of the right forearm sensor. We compute the average and the standard deviation over the
window as features for each axis. We assign the most
frequent label to the window.  The resulting data
points were shuffled uniformly.

In addition, we construct another dataset from \banosdataset, in which we
simulate a concept drift by shifting the activity labels in the
second half of the data stream. This is useful to
observe any behavioral change induced by the
concept drift such as an increase in power
consumption.

\subsubsection{\recofitdataset}
The \recofitdataset dataset~\cite{recofit} is a
human activity dataset containing 94
participants\footnote{\recofitdataset dataset
available
\href{https://msropendata.com/datasets/799c1167-2c8f-44c4-929c-227bf04e2b9a}{here}.}. Similarly to the \banosdataset
dataset, the activity labeled 0 indicates no
specific activity.
Since many of these activities were similar, we
merged  some of them together based on the table
in~\cite{behzad2019}. 

We pre-processed the dataset similarly to the
\banosdataset one, using non-overlapping windows of
one second, and only using 6 axes (acceleration
and gyroscope) from one sensor.
. From these 6 axes, we used the average and the standard deviation
over the window as features. We assigned the most
frequent label to the window.

\subsubsection{MOA dataset}
Massive Online Analysis~\cite{moa} (MOA) is a Java framework to compare
data stream classifiers. In addition to classification algorithms, MOA provides many
tools to read and generate datasets.
We generate three synthetic datasets\footnote{MOA commands available \href{https://github.com/azazel7/paper-benchmark/blob/e0c9a94d0d17490f7ab14293dec20b8322a6447c/Makefile\#L90}{here}.}:
a hyperplane, a RandomRBF, and a RandomTree
dataset. We generate 200,000 data points
 for each of these synthetic datasets.
The hyperplane and the RandomRBF both have three features and two classes, however, the RandomRBF has a slight imbalance toward one class.
The RandomTree dataset is the hardest of the three, with six attributes and
ten classes. Since the data points are generated with a tree structure, we
expect the decision trees to show better performances than the other
classifiers.

\subsection{Algorithms and Implementation}
In this section, we describe the algorithms used in the benchmark, their
hyperparameters, and relevant implementation details. 

\subsubsection{Mondrian forest~(\mondrianforest)~\cite{mondrian2014}}
Each tree in a \mondrianforest recursively splits
the feature space, similar to a regular decision tree.
However, the feature used in the split and the
value of the split are picked randomly. The
probability to select a feature is proportional to
its normalized range, and the value for the split is
uniformly selected in the range of the feature.

In OrpailleCC, the amount of memory allocated to the forest is predefined,
and it is shared by all the trees in the forest, leading to a constant
memory footprint for the classifier. This implementation is memory-bounded,
meaning that the classifier can adjust to memory limitations, for instance
by stopping tree growth or replacing existing nodes with new ones. This is
different from an implementation with a constant space complexity, where
the classifier would use the same amount of memory regardless of the
amount of available memory. For instance, in our study, the \mondrianforest
classifier is memory-bounded while \naivebayes classifier has a constant
space complexity.

Mondrian trees can be tuned using three
parameters: the base count, the discount factor,
and the budget. The base count is used to
initialize the prediction for the root. The
discount factor influences the nodes on how much
they should use their parent prediction. A
discount factor closer to one makes the prediction
of a node closer to the prediction of its parent.
Finally, the budget controls the tree depth.

Hyperparameters used for \mondrianforest are available in the
\href{https://github.com/azazel7/paper-benchmark/blob/master/README.md}{repository readme}.
Additionally, the \mondrianforest is allocated
with 600~KB of memory unless specified otherwise.
On the \banosdataset and \recofitdataset datasets,
we also explore the \mondrianforest with 3~MB of
memory in order to observe the effect of available memory on
performances (classification, runtime, and power).
\subsubsection{Micro Cluster Nearest Neighbor~\cite{mc-nn}}
The Micro Cluster Nearest Neighbor~(\mcnn) is a
variant of k-nearest neighbors where data points
are aggregated into clusters to reduce storage
requirements.  
The space and time complexities of \mcnn are
constant since the maximum number of clusters is fixed.
The reaction to concept drift is influenced by the
participation threshold and the error threshold.
A higher participation threshold and a lower error
threshold increase reaction speed to concept
drift. Since the error thresholds used in this
study are small, we expect \mcnn to react quite
fast and efficiently to concept drifts.

We implemented two versions of \mcnn in
OrpailleCC, differing in the way they remove
clusters during training. The first version (\mcnn
Origin) is similar to the mechanism described
in~\cite{mc-nn}, based on participation scores.
The second version (\mcnn OrpailleCC)
removes the cluster with the lowest participation
only when space is needed.  A cluster slot is
needed when an existing cluster is split and there
is no more slot available because the number of
active clusters already reached the maximum defined
by the user.

\mcnn OrpailleCC has only one
parameter, the error threshold after which a
cluster is split.  \mcnn Origin has two
parameters: the error threshold and the
participation threshold. The participation
threshold is the limit below which a cluster is
removed.
Hyperparameters used for \mcnn are available in the
\href{https://github.com/azazel7/paper-benchmark/blob/master/README.md}{repository readme}.

\subsubsection{Naïve Bayes~(\naivebayes)~\cite{naive_bayes}}
The \naivebayes algorithm keeps a table of
counters for each feature value and each label.
During prediction, the algorithm assigns a
score for each label depending on how the data
point to predict compares to the values observed
during the training phase. 

The implementation from \streamdmcpp was used in this
benchmark. It uses a Gaussian
fit for numerical attributes. Two implementations were used, the OrpailleCC one and the StreamDM one. We used two implementations 
to provide a comparison reference between the two libraries.

\subsubsection{Hoeffding Tree~(\hoeffdingtree)~\cite{VFDT}}
Similar to a decision tree, the \hoeffdingtree recursively splits the feature
space to maximize a metric, often the information gain or the Gini index.
However,  to estimate when a leaf should be split, the \hoeffdingtree relies on
the Hoeffding bound, a measure of the score deviation of the splits. We used
this classifier as implemented in \streamdmcpp.  The \hoeffdingtree is common in
data stream classification, however, the internal nodes are static and cannot be
re-considered. Therefore, any concept drift adaption relies on the new leaves
that will be split.

The \hoeffdingtree has three parameters: the
confidence level, the grace period, and the leaf
learner. The confidence level is the probability that
the Hoeffding bound makes a wrong estimation of
the deviation. The grace period is the number of
processed data points before a leaf is evaluated for a split.
 The leaf learner is the method used in the
leaf to predict the label.  In this study, we used
a confidence level of $0.01$ with a grace period
of 10 data points and the \naivebayes classifier as leaf
learner.

\subsubsection{Feedforward Neural Network~(\FNN)}
\label{sec:method-fnn}
A neural network is a combination of artificial neurons, also known as
perceptrons, that all have input weights and an activation function. In this
benchmark, we used a fully-connected \FNN, that is, a network where perceptrons
are organized in layers and all output values from perceptrons of layer $n-1$
serve as input values for perceptrons of layer $n$.  We used a 3-layer network
with 120 inputs, 30 perceptrons in the hidden layer, and 33 output perceptrons.
Because a \FNN takes many epochs to update and converge it barely adapts to a
concept drifts even though it trains with each new data point.

In this study, we used histogram features
from~\cite{omid_2019} instead of the ones
presented in Section~\ref{sec:method-dataset}
because the network performed
poorly with these features. The histogram features
produce 20 bins per axis.

This neural network can be tuned by changing the
number of layers and the size of each layer.
Additionally, the activation function and the
learning ratio can be changed. The learning ratio
indicates by how much the weights should change
during backpropagation.

\subsubsection{Hyperparameters Tuning}
For each classifier, we tuned hyperparameters  using the first subject from the
\banosdataset dataset.  The data from this subject was pre-processed as the rest
of the \banosdataset dataset (window size of one second, average and standard
deviation on the six-axis of the right forearm sensor, $\ldots$). We did a grid
search to test multiple values for the parameters.

The classifiers start the prequential phase with
no knowledge from the first subject.  We made an
exception for the \FNN because we noticed that it
performed poorly with random weights and it needed
many epochs to achieve better performances than a
random  classifier. Therefore, we decided to
pre-train the \FNN and re-use the weights as a
starting point for the prequential phase.

For other classifiers, only the 
hyperparameters were taken from the tuning phase.
We selected the hyperparameters that 
maximized the F1 score on the first subject.

\subsubsection{Offline Comparison}
We compared data stream algorithms with an offline \knn. The value of $k$ were
selected using a grid search.

\subsection{Evaluation}
We computed four metrics: the F1 score, the memory
footprint, the runtime, and the power usage.
The F1 score and the memory
footprint were computed periodically during the
execution of a classifier. The
power consumption and the runtime were collected
at the end of each execution.

\paragraph{Classification Performance}
We monitor the true positives, false positives,
true negatives, and false negatives using the
prequential evaluation, meaning that with each new
data point the model is first tested and then
trained.  From these counts, we compute the
F1 score every 50 elements. We do not apply any
fading factor to attenuate errors throughout the
stream.  We compute the F1 score in a
one-versus-all fashion for each class, averaged
across all classes (macro-average, code available
\href{https://github.com/azazel7/paper-benchmark/blob/9adb1039c5a65a00a66d554f0e870d14d3fff7cb/main.cpp\#L82}{here}).
When a class has not been encountered yet, its
F1 score is ignored. We use the F1 score rather
than the accuracy because the real data sets are
imbalanced.

\paragraph{Memory}
We measure the memory footprint by reading file
\texttt{/proc/self/statm} every 50 data points.

\paragraph{Runtime}
The runtime of a classifier is the time needed for
the classifier to process the dataset. We
collect the runtime reported by the \textit{perf}
command\footnote{\textit{perf}
\href{https://perf.wiki.kernel.org/index.php/Main_Page}{website}}, which
includes loading of the binary in memory, setting up data structures, and
opening the dataset file. To remove these overheads from our measurements,
we use the runtime of an empty classifier that always predict class 0 as a baseline.

\paragraph{Power} We measure the average power consumed by classification
algorithms with the \textit{perf} command. The power measurement is done
multiple times in a minimal environment. We use the empty classifier as a
baseline.

\subsection{Experimental Conditions}
We automated our experiments with a Python script that defines
classifiers and their parameters, randomizes all
the repetitions, and plots the
resulting data. The datasets and output results were stored in memory
through a memfs filesystem mounted on \texttt{/tmp}, to reduce the impact of I/O time.
We averaged metrics across repetitions (same classifier, same parameters, and
same dataset).

The experiment was done with a single core of a
cluster node with two Intel(R) Xeon(R)
Gold 6130 CPUs and a main memory of 250G.


\begin{figure*}
	\centering
	\includegraphics[width=0.8\linewidth]{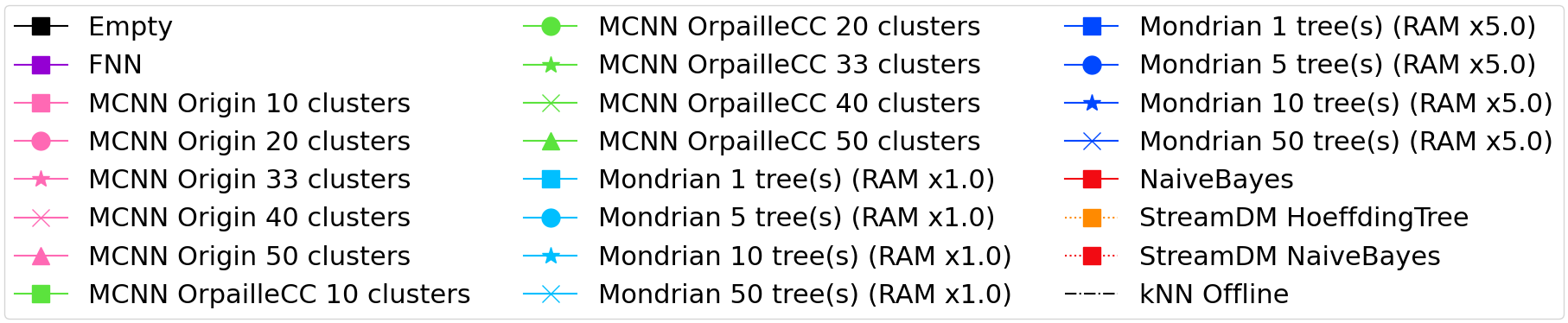}
	\begin{subfigure}[t]{.49\linewidth}
		\includegraphics[width=\linewidth]{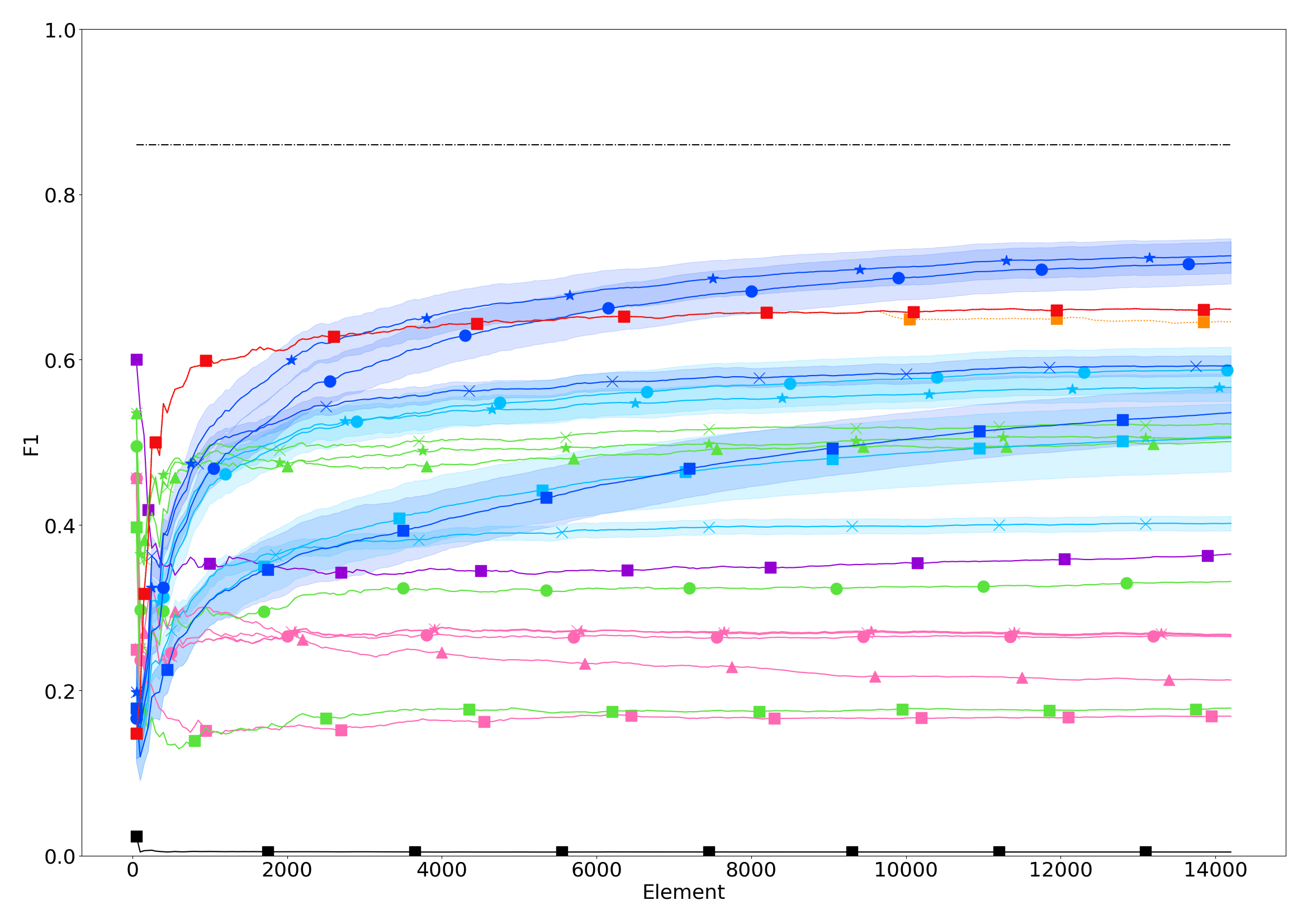}
		\caption{\banosdataset}
		\label{fig:f1-banos}
	\end{subfigure}
	\begin{subfigure}[t]{.49\linewidth}
		\includegraphics[width=\linewidth]{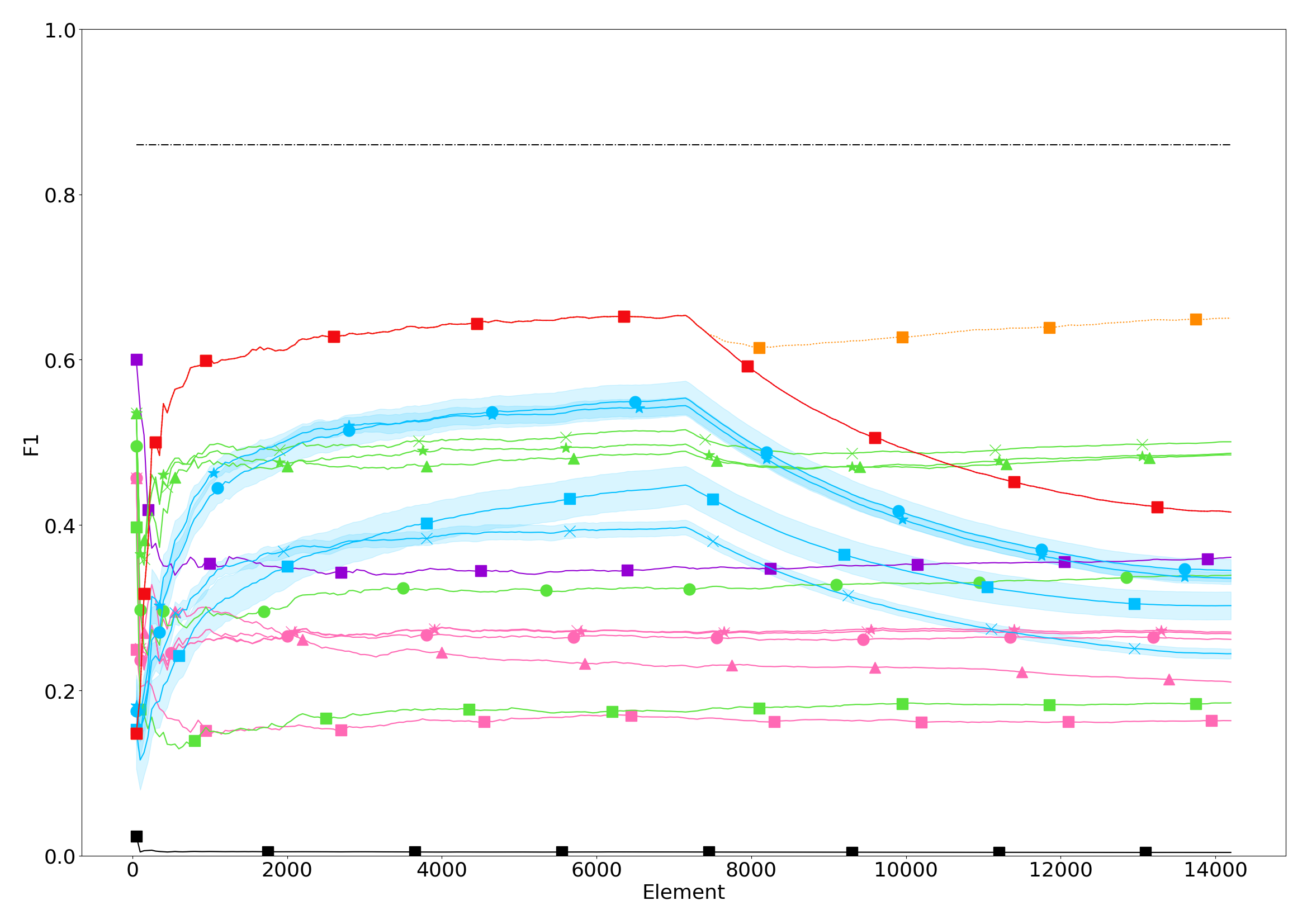}
		\caption{\banosdataset (with Drift)}
		\label{fig:f1-drift}
	\end{subfigure}\\
	\begin{subfigure}[t]{.49\linewidth}
		\includegraphics[width=\linewidth]{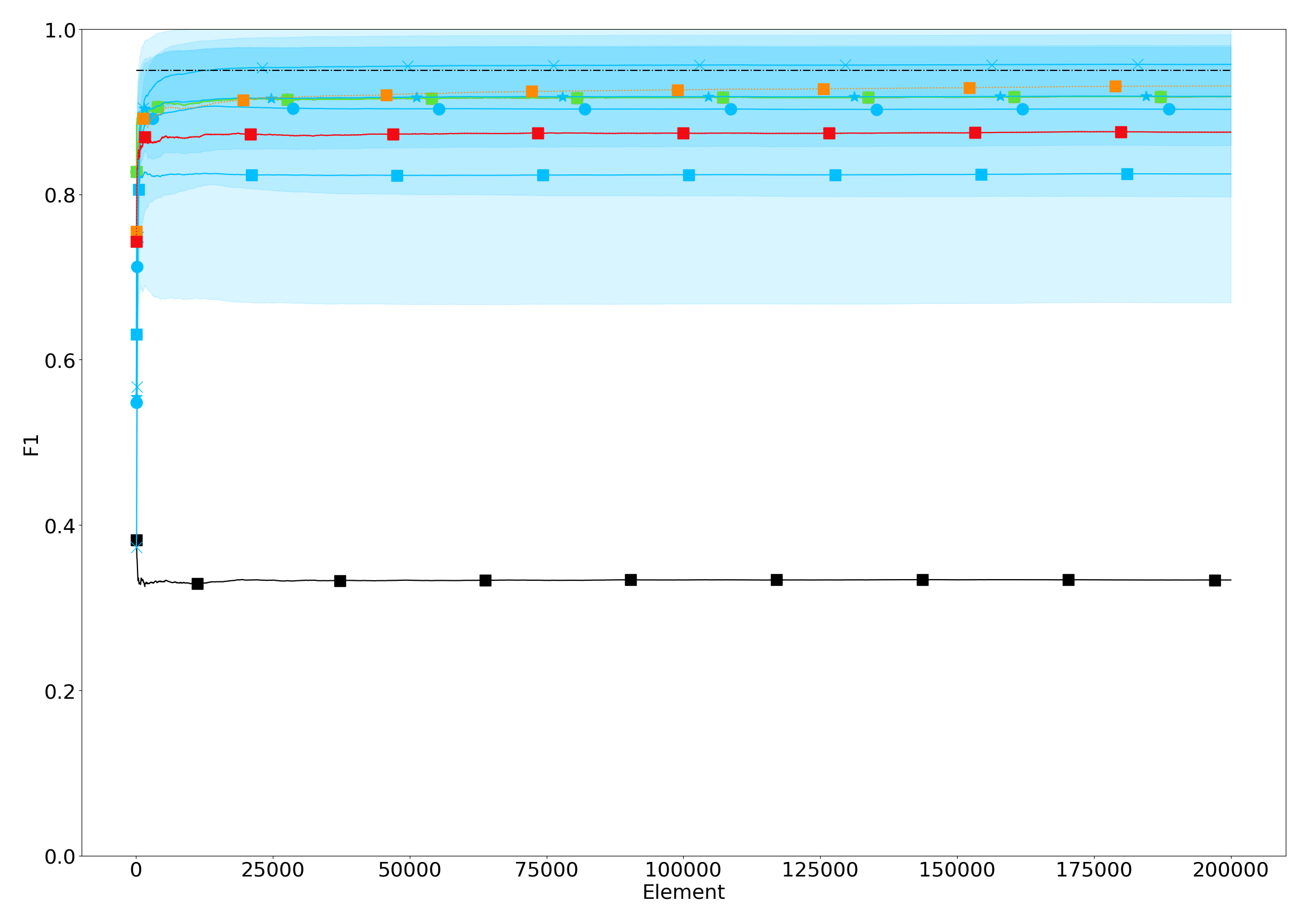}
		\caption{Hyperplane (MOA)}
		\label{fig:f1-dataset_1}
	\end{subfigure}
	\begin{subfigure}[t]{.49\linewidth}
		\includegraphics[width=\linewidth]{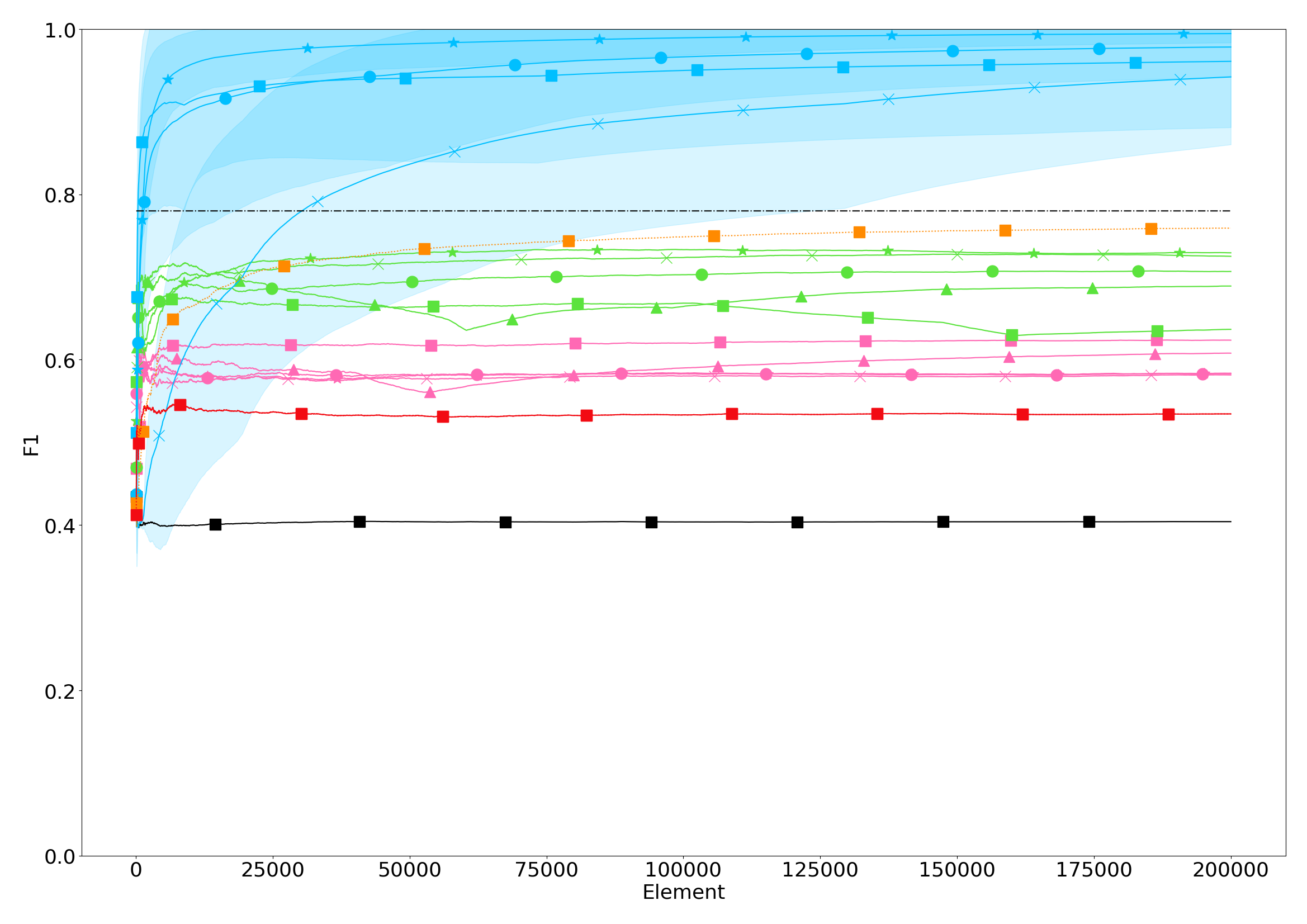}
		\caption{RandomRBF (MOA)}
		\label{fig:f1-dataset_2}
	\end{subfigure}\\
	\begin{subfigure}[t]{.49\linewidth}
		\includegraphics[width=\linewidth]{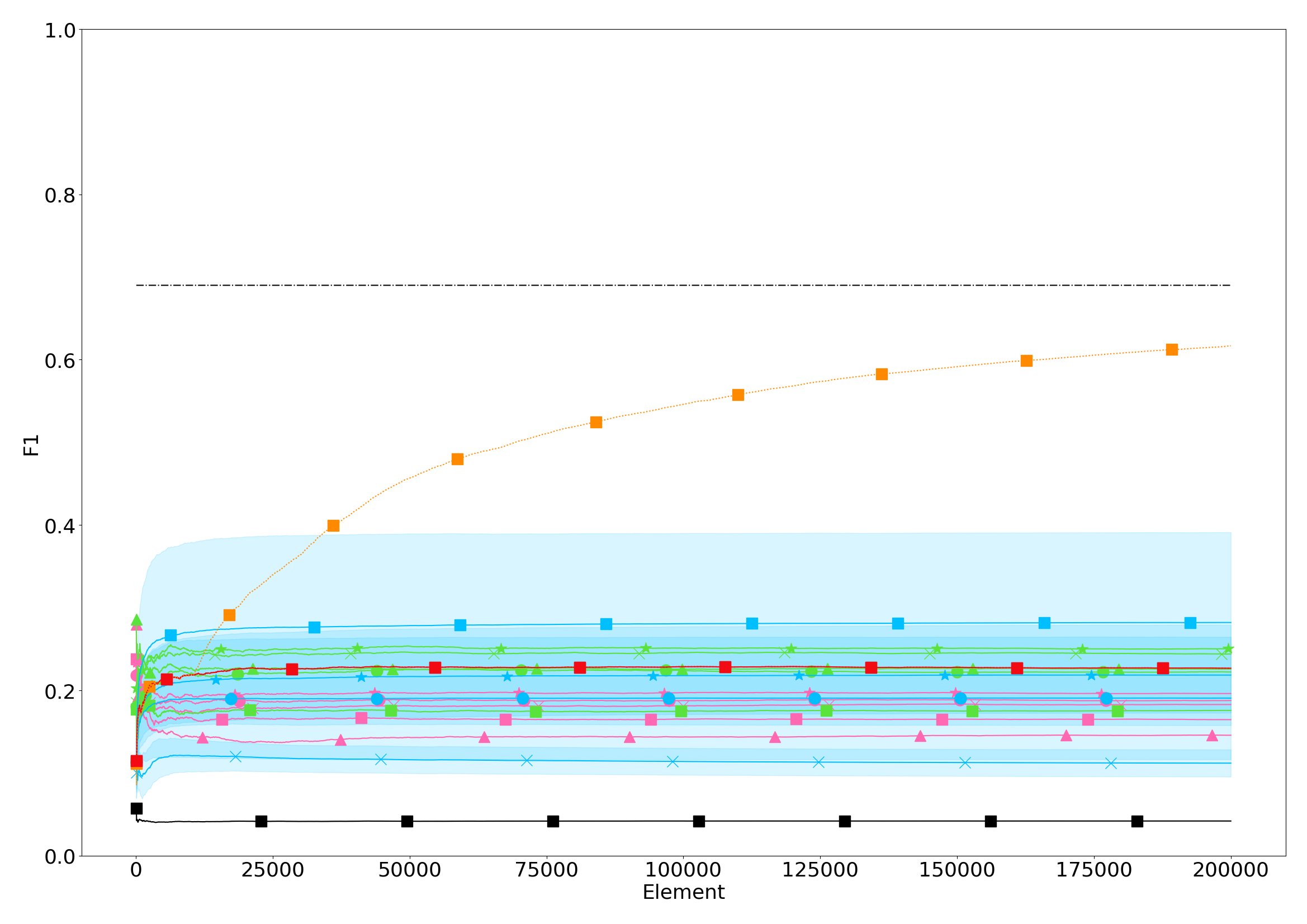}
		\caption{RandomTree (MOA)}
		\label{fig:f1-dataset_3}
	\end{subfigure}
	\begin{subfigure}[t]{.49\linewidth}
		\includegraphics[width=\linewidth]{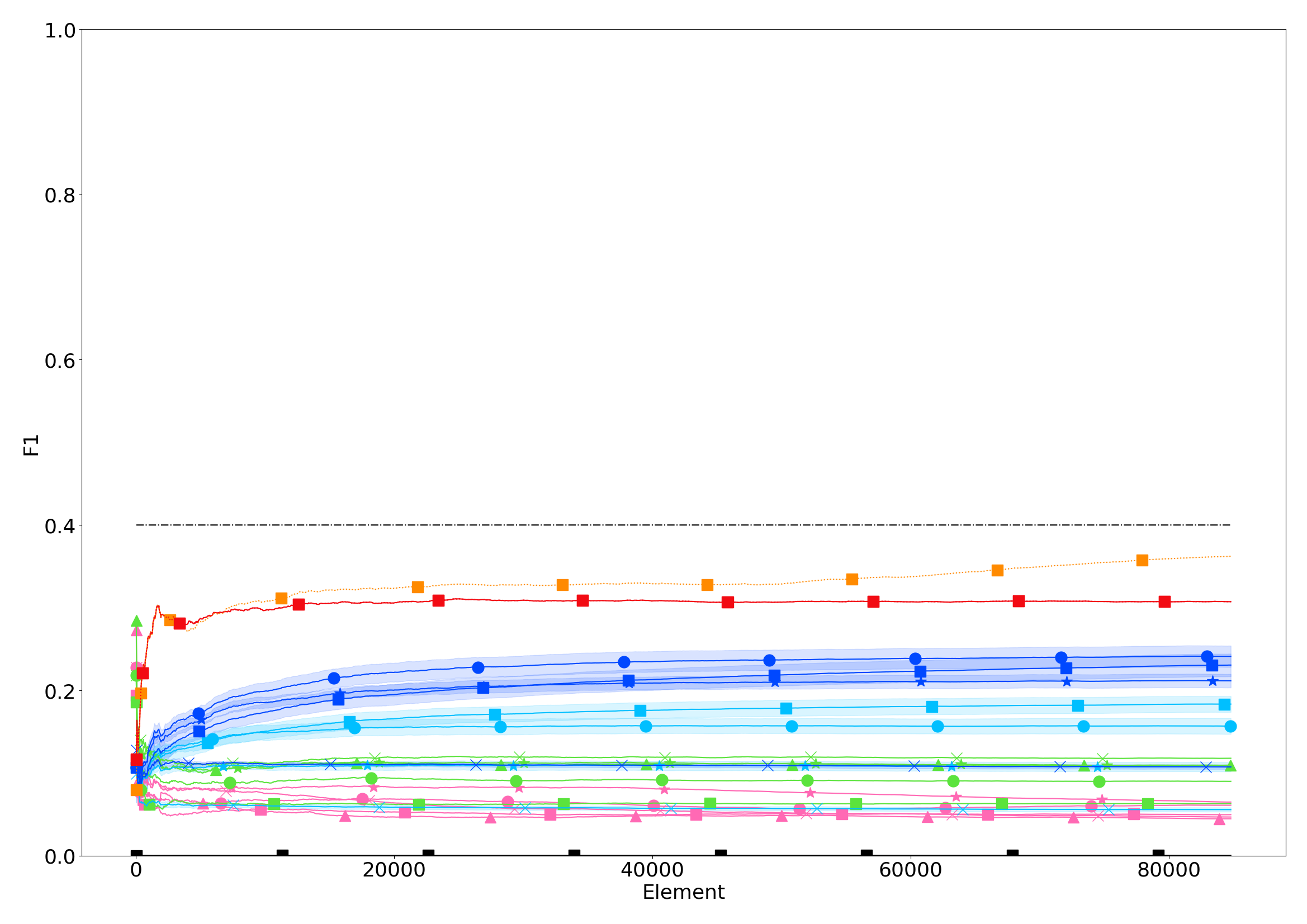}
		\caption{\recofitdataset}
		\label{fig:f1-recofit}
	\end{subfigure}
	\caption{F1 scores for the six datasets (average over 20 repetitions).
		The horizontal dashed black line indicates the offline \knn F1
		score (the value of k was obtained by grid search in [2, 20]). The
		blue shades represent a $\pm\sigma$ interval of the \mondrianforest
		classifier across repetitions.}
	\label{fig:f1}
\end{figure*}

\section{Results}
This section presents our benchmark results and the corresponding
hyperparameter tunning experiments.

\subsection{Overall classification performance}

Figure~\ref{fig:f1} compares the F1 scores obtained by all classifiers on
the six datasets. The graphs also show the standard deviation of the
\mondrianforest classifier observed across all repetitions (the other
classifiers do not involve any randomness). 

F1 scores vary greatly across the datasets. While the highest
observed F1 score is above 0.95 on the Hyperplane and RandomRBF datasets,
it barely reaches 0.65 for the \banosdataset dataset, and it remains under
0.40 on the \recofitdataset and RandomTree datasets. This trend is
consistent for all classifiers.

The offline \knn classifier used as baseline achieves better F1 scores than
all other classifiers, except for the \mondrianforest on the Hyperplane and
the RandomRBF datasets. On the \banosdataset dataset, the difference of
0.23 with the best stream classifier remains very substantial, which
quantifies the remaining performance gap between data stream and offline
classifiers. On the \recofitdataset dataset, the difference between stream
and offline classifiers is lower, but the offline performance remains very
low.

It should be noted that the F1 scores observed for the offline \knn
classifier on the real datasets are substantially lower than the values
reported in the literature. On the \banosdataset dataset, the original
study in~\cite{Banos_2014} reports an F1 score of 0.96, the work
in~\cite{behzad2019} achieves 0.92, but our benchmark only achieves 0.86.
Similarly, on the \recofitdataset dataset, the original study reports an
accuracy of 0.99 and the
work in~\cite{behzad2019} reaches 0.65 while our benchmark only achieves
0.40. This is most likely due to our use of data coming from a single
sensor, consistently with our motivating use case, while the other works used
multiple ones (9 in the case of
\banosdataset).


The \hoeffdingtree appears to be the most robust to concept drifts
(\banosdataset with drift), while the \mondrianforest and \naivebayes
classifiers are the most impacted. \mcnn classifiers are marginally impacted.
The low resilience of \mondrianforest to concept drifts can be attributed to
two factors. First, existing nodes in trees of a \mondrianforest cannot be updated.
Second, when the memory limit is reached, \mondriantrees cannot grow
or reshape their structure anymore.

\subsection{\hoeffdingtree and \naivebayes}

The \naivebayes and the \hoeffdingtree classifiers stand out on the two real datasets
(\banosdataset and \recofitdataset) even though the F1 scores observed remain
low (0.6 and 0.35) compared to offline \knn (0.86 and 0.40). Additionally, the
\hoeffdingtree performs outstandingly on the RandomTree dataset and
\banosdataset dataset with a drift. Such good performances were expected on the
RandomTree dataset because it was generated based on a tree structure.

Except for the \banosdataset dataset, the \hoeffdingtree performs better
than \naivebayes. For all datasets, the performance of both classifiers is
comparable at the beginning of the stream, because the \hoeffdingtree uses
a \naivebayes in its leaves.  However, F1 scores diverge throughout the
stream, most likely because of the \hoeffdingtree's ability to reshape its
tree structure.  This occurs after a sufficient amount of elements, and the
difference is more noticeable after a concept drift.

Finally, we note that the \streamdmcpp and OrpailleCC implementations of
\naivebayes are indistinguishable from each other, which confirms the correctness of our
implementation in OrpailleCC.

\subsection{\mondrianforest}

On two synthetic datasets, Hyperplane and RandomRBF, the \mondrianforest (RAM x
1.0) with 10 trees achieves the best performance (F1$>$0.95), above offline
\knn.  Additionally, the \mondrianforest with 5 or 10 trees ranks third on the
two real datasets. 

Surprisingly, a \mondrianforest with 50 trees performs worse than 5 or 10
trees on most datasets. The only exception is the Hyperplane dataset where
50 trees perform between 5 and 10 trees. This is due to the fact that
our \mondrianforest implementation is memory-bounded, which is
useful on connected objects but limits tree growth when the allocated memory is
full. Because 50 trees fill the memory faster than 10 or 5 trees, the
learning stops earlier, before the trees can learn enough from the
data. It can also be noted that the variance of the F1 score decreases with
the number of trees, as expected.

The dependency of the \mondrianforest to memory allocation is shown in
\banosdataset and \recofitdataset datasets, where an additional
configuration with five times more memory than the initial configuration was run
 (total of 3~MB).  The memory increase induces an F1 score difference
greater than 0.1, except when only one tree is used, in
which case the improvement caused by the memory is less than 0.05. Naturally, the
selected memory bound may not be achievable on a connected object. Overall,
\mondrianforest seems to be a viable alternative to \naivebayes or the
\hoeffdingtree for \har.

\subsection{\mcnn}

The \mcnn OrpailleCC stands out on the \banosdataset (with drift) dataset where it
ranks second thanks to its adaptation to the concept drift.  On other datasets,
\mcnn OrpailleCC ranks below the \mondrianforest and the \hoeffdingtree, but
above \mcnn Original. This difference between the two \mcnn implementations is presumably due
to the fact that \mcnn Origin removes clusters with low participation too early.
On the real datasets (\banosdataset and \recofitdataset), we notice that the
\mcnn OrpailleCC appears to be learning faster than the \mondrianforest,
although the \mondrianforest catches up after a few thousand elements. Finally,
we note that \mcnn remains quite lower than the offline \knn.

\subsection{\FNN}

Figure~\ref{fig:f1-banos} shows that the \FNN has a low F1 score (0.36)
compared to other classifiers (above 0.5), which contradicts the results
reported in~\cite{omid_2019} where the \FNN achieves more than 95\%
accuracy in a context of offline training. The main difference
between~\cite{omid_2019} and our study lies in the definition of the
training set. In~\cite{omid_2019}, the training set includes examples from
every subject, while we only use a single one, to ensure an objective
comparison with the other stream classifiers that do not require offline
training (except for hyperparameter tuning, done on the first subject of
the \banosdataset dataset). When we use a random sample of 10\% of the
datapoints across all subjects for offline training, we reach an F1 score
of 0.68, which is higher than the performance of the \naivebayes classifier.


\begin{figure}
		\includegraphics[width=\linewidth]{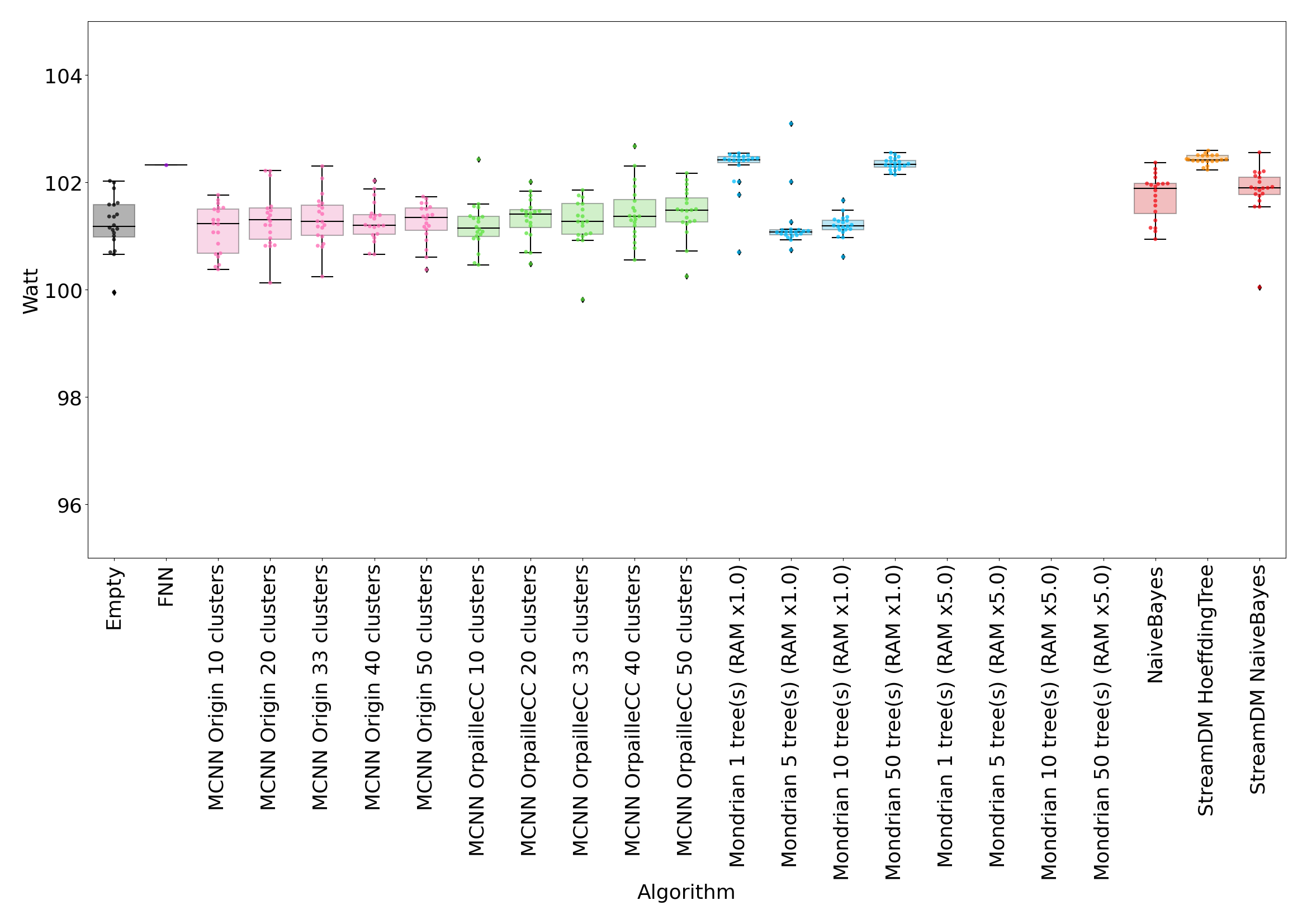}
		\caption{Power usage for the \banosdataset dataset. Results are similar across
		datasets.}
		\label{fig:power}
\end{figure}
\subsection{Power}
\label{sec:result-power}
Figure~\ref{fig:power} shows the power usage of each classifier on four
datasets (results are similar for the other two datasets). All classifiers exhibit comparable power consumptions, close to
102~W. 

This observation is explained by two factors. First, the benchmarking
platform was  working at minimal power. To ensure no disturbance by a
background process, we run the classifiers on an isolated cluster node with
eight cores. Therefore, the power difference on one core is not noticeable.

Another reason is the dataset sizes. Indeed, the slowest run is about
10 seconds with 50 Mondrian trees on \recofitdataset dataset. Such short
executions do not leave time for the CPU to switch P-states because it
barely warms a core. Further experiments would be required to check how 
our power consumption observations generalize to 
connected objects. 

\subsection{Runtime}

\begin{figure}
	\includegraphics[width=\linewidth]{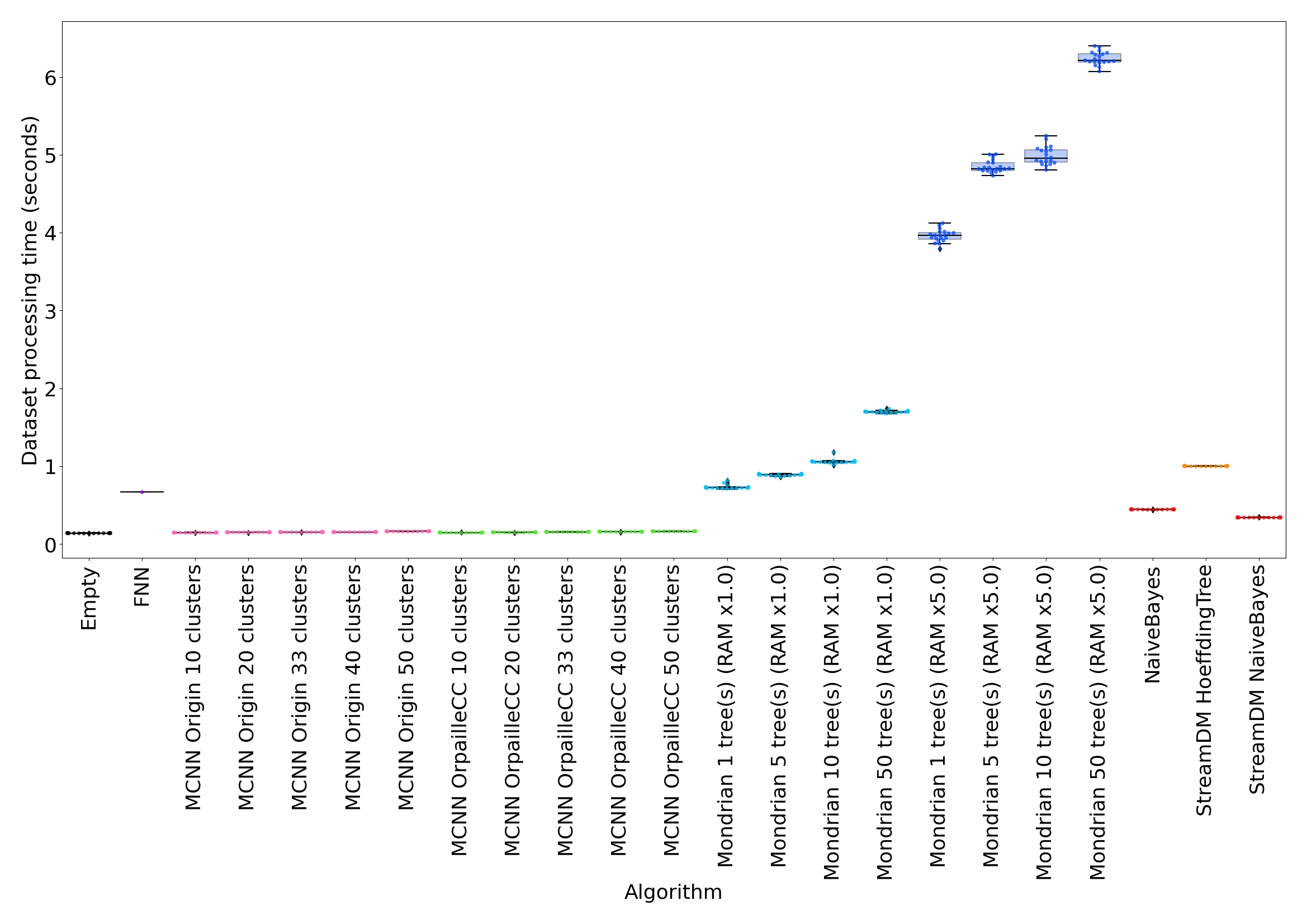}
	\caption{Runtime with the \banosdataset dataset (20 repetitions). Runtime
		results are similar across datasets.}
	\label{fig:runtime}
\end{figure}

Figure~\ref{fig:runtime} shows the classifier runtimes for the two real
datasets. The \mondrianforest is the slowest classifier, in particular for 50
trees which reaches 2 seconds on \banosdataset dataset. This represents roughly
0.35~ms/element with a slower CPU. The second slowest classifier is the
\hoeffdingtree, with a runtime comparable to the \mondrianforest with 10 trees.
The \hoeffdingtree is followed by the two \naivebayes implementations, which is
not surprising since \naivebayes classifiers are used in the leaves of the
\hoeffdingtree. The \mcnn classifiers are the fastest ones, with a runtime very
close to the empty classifier. Note that allocating more memory to the
\mondrianforest substantially increases runtime.

We observe that the runtime of \streamdmcpp's \naivebayes is comparable to
OrpailleCC's. This suggests that the performance of the two libraries is
similar, which justifies our comparison of \hoeffdingtree and \mondrianforest.

\subsection{Memory}
\label{sec:result-memory}
\begin{figure}
	\includegraphics[width=\linewidth]{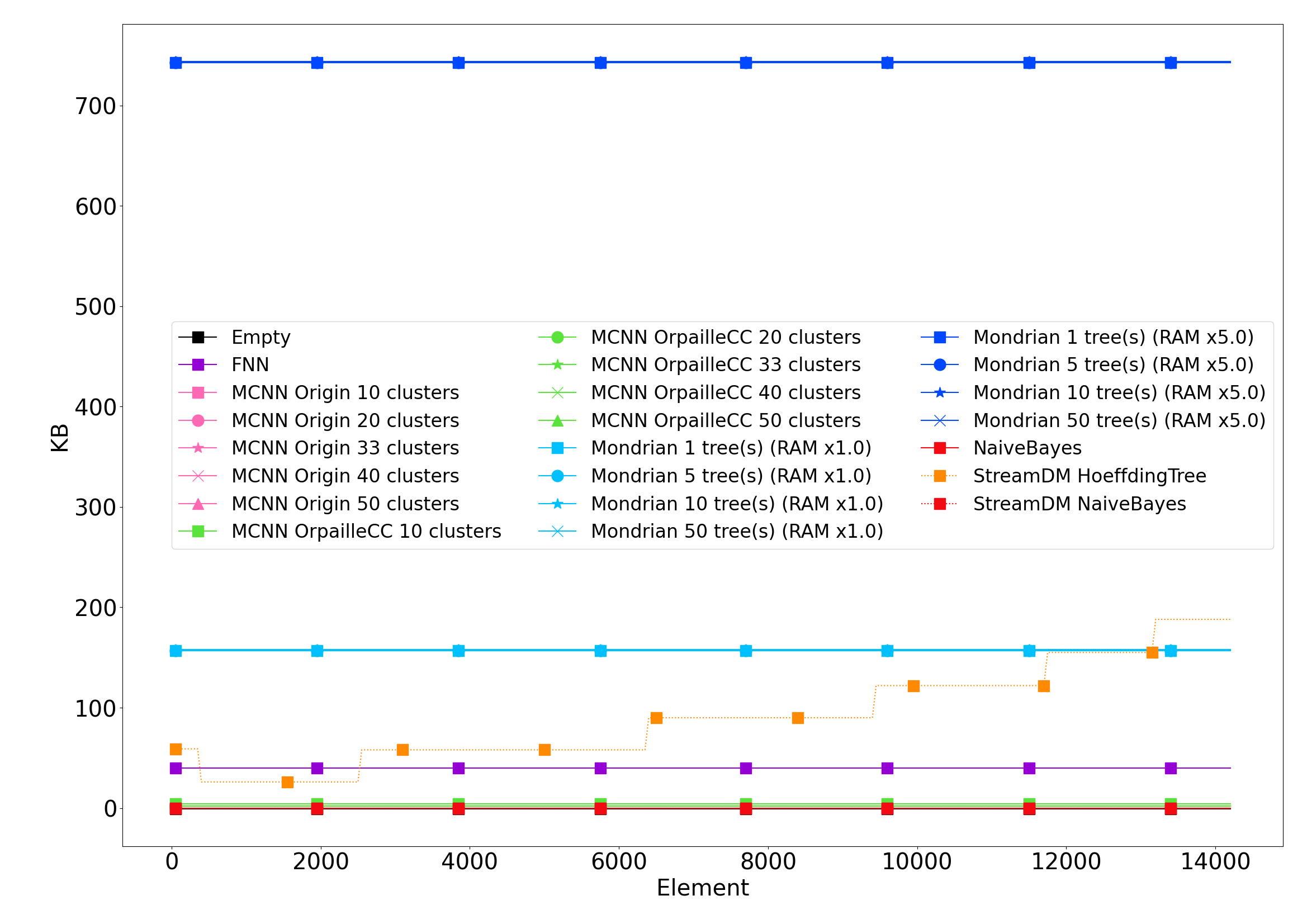}
	\caption{Memory footprint of classifiers with the empty
	classifier as a baseline, measured on the \banosdataset dataset. The memory footprint of the empty
	classifier is 3.44~MB. The baselines are the two \naivebayes from OrpailleCC
		and \streamdmcpp. Their respective memory footprints are 3.44~MB and
		4.74~MB.}
	\label{fig:memory}
\end{figure}

Figure~\ref{fig:memory} shows the evolution of the memory footprint for the
\banosdataset dataset. Results are similar for the other datasets and are
not reported for brevity. Since the  memory footprint of the \naivebayes
classifier was almost indistinguishable from the empty classifier, we used
the two \naivebayes as a baseline for the two libraries. This enables us to
remove the 1.2~MB overhead induced by \streamdmcpp. The \streamdmcpp memory
footprint matches the result in~\cite{StreamDM-CPP}, where the
\hoeffdingtree shows a memory footprint of 4.8~MB.

We observe that the memory footprints of the \mondrianforest and the
\hoeffdingtree are substantially higher than for the other classifiers, which makes 
their deployment on connected objects challenging.
Overall, memory footprints are similar across datasets, due to the
fact that most algorithms follow a bounded memory policy or have a constant
space complexity.  The only exception is the \hoeffdingtree that constantly
selects new splits depending on new data points. The \mondrianforest has the
same behavior but the OrpailleCC implementation is memory-bounded, which
makes its memory footprint constant.

\section{Conclusion}

We conclude that the \hoeffdingtree, the \mondrianforest, and the
\naivebayes data stream classifiers have an overall superiority over the
\FNN and the \mcnns ones for \har.  However, the prediction performance
remains quite low compared to an offline \knn classifier, and it varies
substantially between datasets. Noticeably, the \hoeffdingtree and the
\mcnns classifiers are more resilient to concept drift that the other ones.

Regarding memory consumption, only the \mcnn and \naivebayes classifiers
were found to have a negligible memory footprint, in the order of a few
kilobytes, which is compatible with connected objects. Conversely, the
memory consumed by a \mondrianforest, a \FNN or a \hoeffdingtree is in the
order of 100~kB, which would only be available on some connected objects.
In addition, the classification performance of a \mondrianforest is
strongly modulated by the amount of memory allocated. With enough memory, a
\mondrianforest is likely to match or exceed the performance of the
\hoeffdingtree and \naivebayes classifiers.

The amount of energy consumed by classifiers is mostly impacted by their
runtime, as all power consumptions were found comparable. The
\hoeffdingtree and \mondrianforest are substantially slower than the other
classifiers, with runtimes in the order of 0.35~ms/element, a performance not compatible 
with common sampling frequencies of wearable sensors. 

Future research will focus on reducing the memory requirements and runtime
of the \hoeffdingtree and the \mondrianforest classifiers. In addition to
improving the deployability of these classifiers on connected objects, this
would also potentially improve their classification performance, since
memory remains a bottleneck in the \mondrianforest.

\section*{Acknowledgement}
This work was funded by a Strategic Project Grant of the Natural Sciences
and Engineering Research Council of Canada. The computing platform was
obtained with funding from the Canada Foundation for Innovation.


\bibliographystyle{plain}
\bibliography{paper}
\end{document}